\documentclass[10pt, conference, letterpaper]{IEEEtran}
\usepackage{cite}
\usepackage[caption=false,font=footnotesize]{subfig}
\usepackage{url}
\usepackage[utf8]{inputenc} 
\usepackage[USenglish,UKenglish,english]{babel} 
\usepackage[T1]{fontenc}    
\makeatletter
\newcommand{\upquotetype}{}
\newcommand{\upquote@aux}[1]{\text{\upquotetype}#1\text{\upquotetype}}
\newcommand{\upquotesingle}{\renewcommand{\upquotetype}{\textquotesingle}\upquote@aux}
\newcommand{\upquotedouble}{\renewcommand{\upquotetype}{\textquotedbl}\upquote@aux}

\usepackage{amsthm}

\usepackage{textcomp}

\usepackage{graphicx} 
\usepackage{epstopdf}
\usepackage{amsfonts}
\usepackage{amsmath}

\usepackage{algorithm}
\usepackage{algorithmic}

\begin{document}
\title{Topical Behavior Prediction from Massive Logs}

\author{\IEEEauthorblockN{Shih-Chieh Su\IEEEauthorrefmark{1}
}
\IEEEauthorblockA{
Qualcomm Inc.\\
San Diego, CA, 92121\\
Email: \IEEEauthorrefmark{1}shihchie@qualcomm.com}}

\maketitle

\begin{abstract}
In this paper, we study the topical behavior in a large scale. We use the network logs where each entry contains the entity ID, the timestamp, and the meta data about the activity. Both the temporal and the spatial relationships of the behavior are explored with the deep learning architectures combing the recurrent neural network (RNN) and the convolutional neural network (CNN). To make the behavioral data appropriate for the spatial learning in the CNN, we propose several reduction steps to form the topical metrics and to place them homogeneously like pixels in the images. The experimental result shows both temporal and spatial gains when compared against a multilayer perceptron (MLP) network. A new learning framework called the spatially connected convolutional networks (SCCN) is introduced to predict the topical metrics more efficiently.
\end{abstract}

\IEEEpeerreviewmaketitle

\section{Introduction}\label{sec-intro}

Understanding and predicting the behavior of an entity over a large domain of different actions is a challenging problem. The problem is even more difficult when the behavioral data is massively collected with lots of noise. For example, the network activity log from nodes within a network domain, the check-ins from users of a social media site, the visited geo-locations from players within a augmented reality game, or the shipment destinations of all items sold at an ecommerce site. Among the data, each entity behaves differently, over different periods of time. What can we learn from the behavioral data?

There are various studies in using behavioral data as a global indicator. For instance, large scale user activity data from Google is used to measure and track the user experience such as happiness and engagement \cite{rodden10}. The web behavioral data including searches and page views is used by Microsoft to decide the advertisement delivered to the user \cite{chandramouli12}. Similarly, Yahoo also conducts study on how education and other factors can affect the web browsing behavior, which can also be applied to improve advertisement targeting \cite{goel12}. The predictor to track stock index can be composed from the categorized moods based on the overall Twitter activities \cite{bollen11}. It is also possible to aim on lots of different business intelligence targets with the behavioral data at hand \cite{chen12}.

However, the aforementioned large scale behavioral analytics use cases have one aspect in common: they heavily simplified the response domain to have one or few learnable targets. If we use the behavior domain as the response domain, can we answer the "what's next" question? For example, if a user checked in at a Starbucks store this month and six months ago, will he checked in at a Starbucks store next month? We may not have enough historical data about this user. But the data from other users can help.

In this paper, we attempt to tackle the "what's next" problem using the historical behavioral data, not only from the target entity itself, but also from the peer entities. First, we organize the activities into topics. The number of topics is chosen according to the desirable size and granularity of the response domain. The topical activities on each topic is then quantified and measured for each entity. Over several periods of time, we observe the topical behavior over the same set of topics for all entities in the experiment. The historical topical activities is used to train the models that predict topical activities in the next time period.

Several combinations of deep neural network (DNN) are explored to predict topical behavior. Specifically, the long short-term memory units (LSTM) \cite{hochreiter97} and other types of RNN \cite{funahashi93} are employed to learn the temporal variation patterns of the topical behavior. The CNN \cite{lawrence97} and the locally-connected network (LCN) \cite{lecun98} are used to learn the spatial composition of the topical behavior. The relationship between topics needs to be abstracted and evenly distributed like pixels for the CNN and the LCN to learn \cite{Su16}. The experiment result is compared against the benchmark result from the multilayer perceptron (MLP) that does not exploit either temporal or spatial relationship.

\begin{figure*}[tb]
  \centering
  \subfloat[]{\includegraphics[scale=0.24]{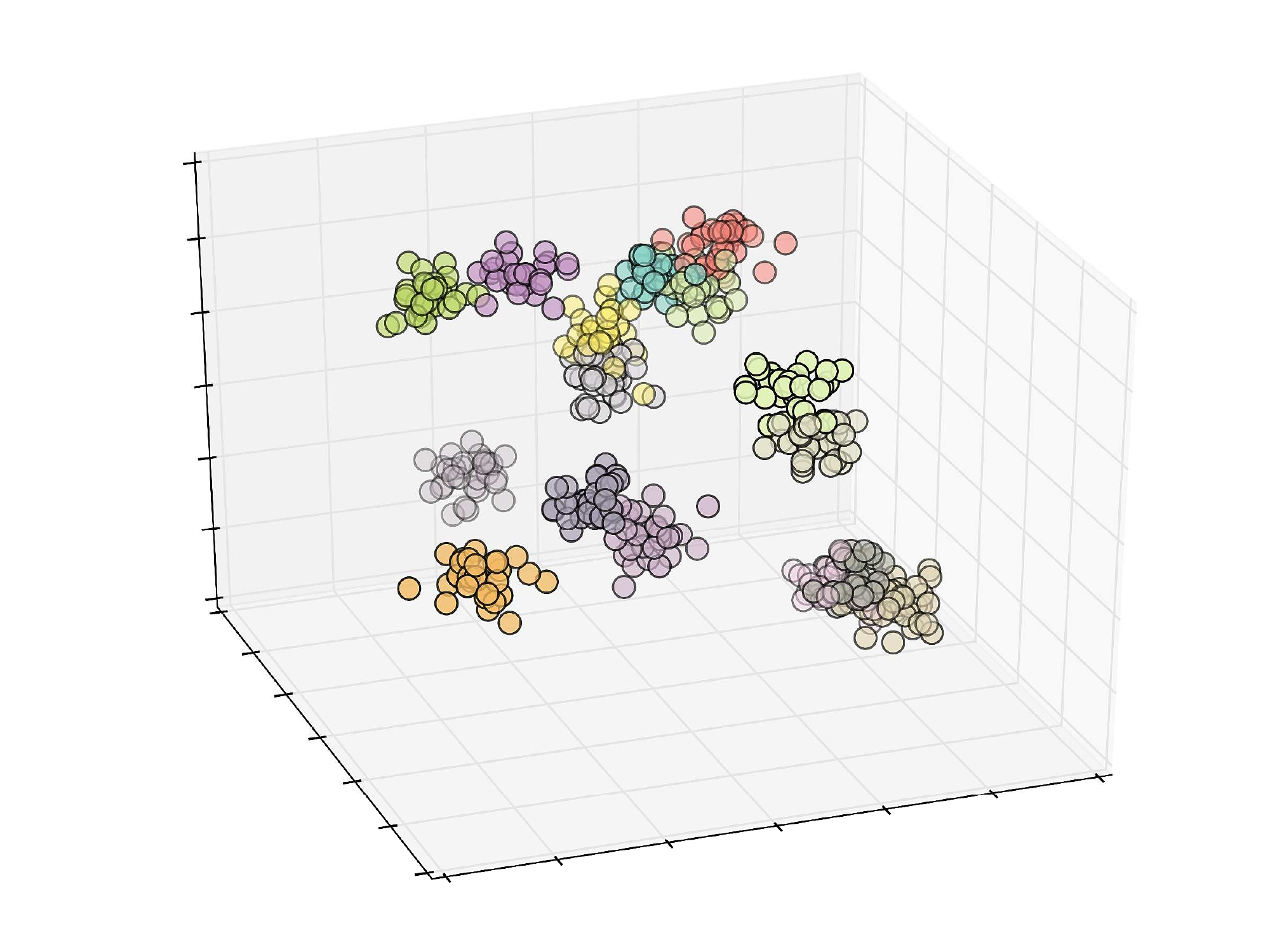}} 
  \subfloat[]{\includegraphics[scale=0.23]{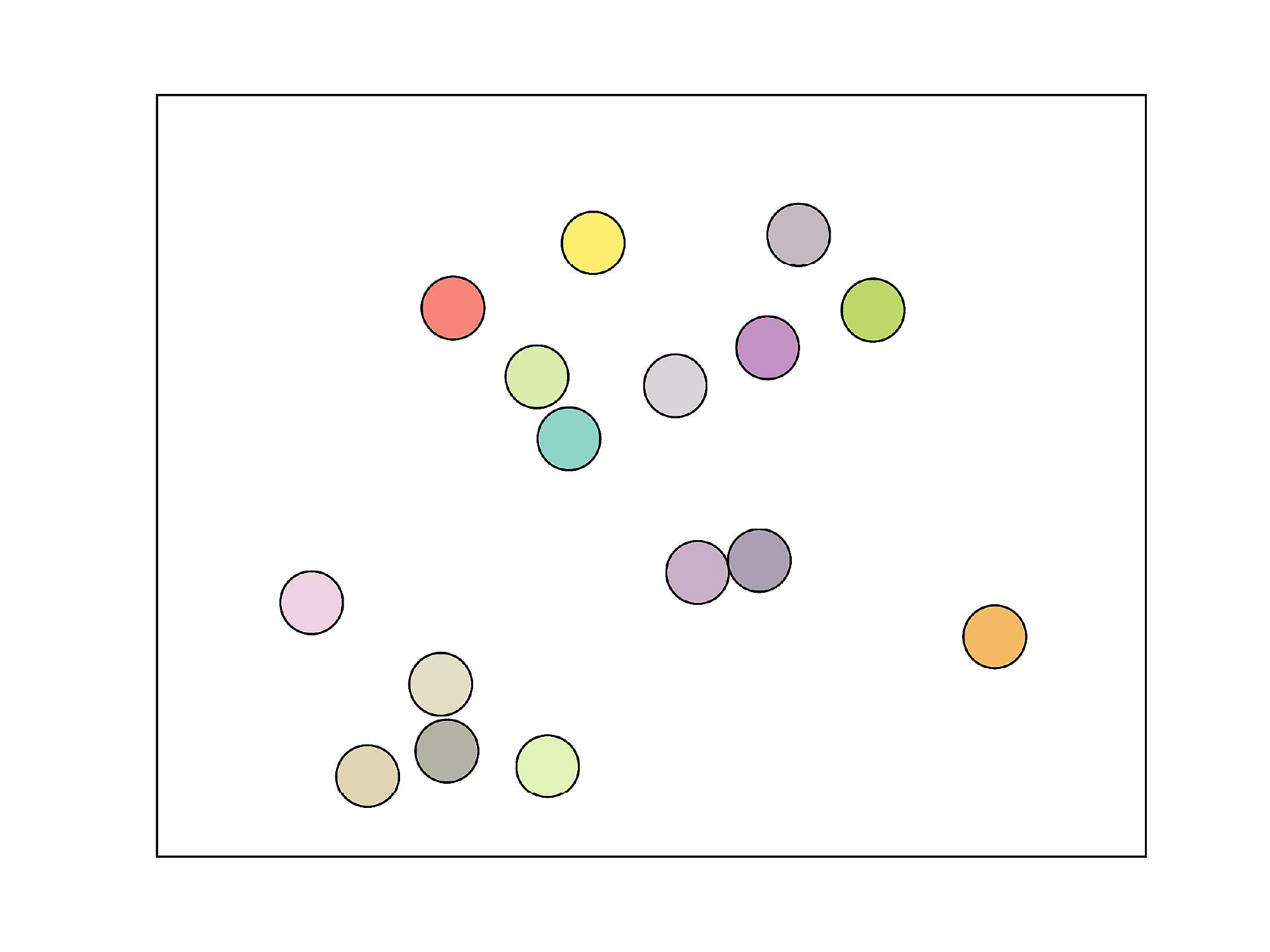}}
  \subfloat[]{\includegraphics[scale=0.23]{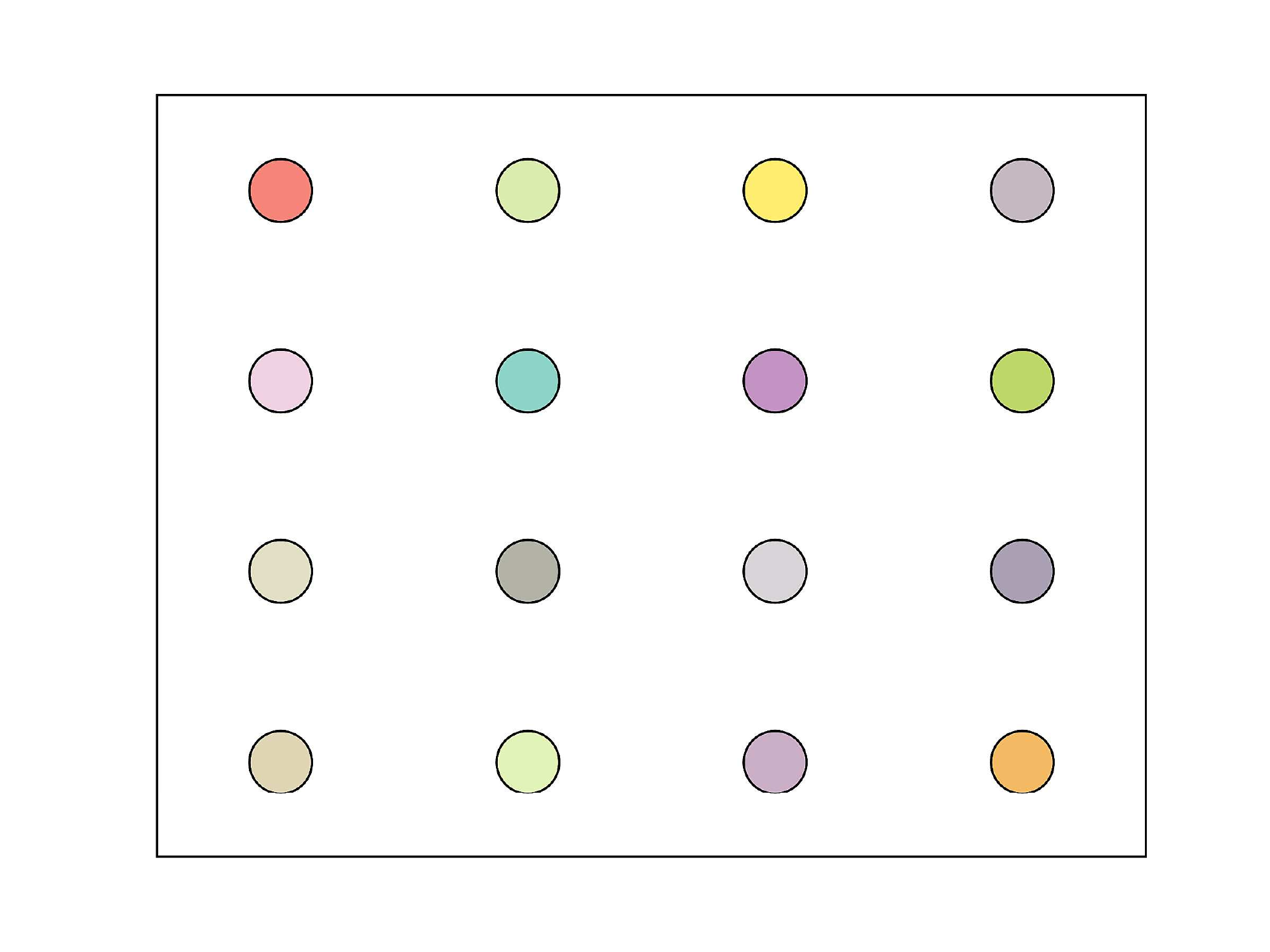}}
  \subfloat[]{\includegraphics[scale=0.036]{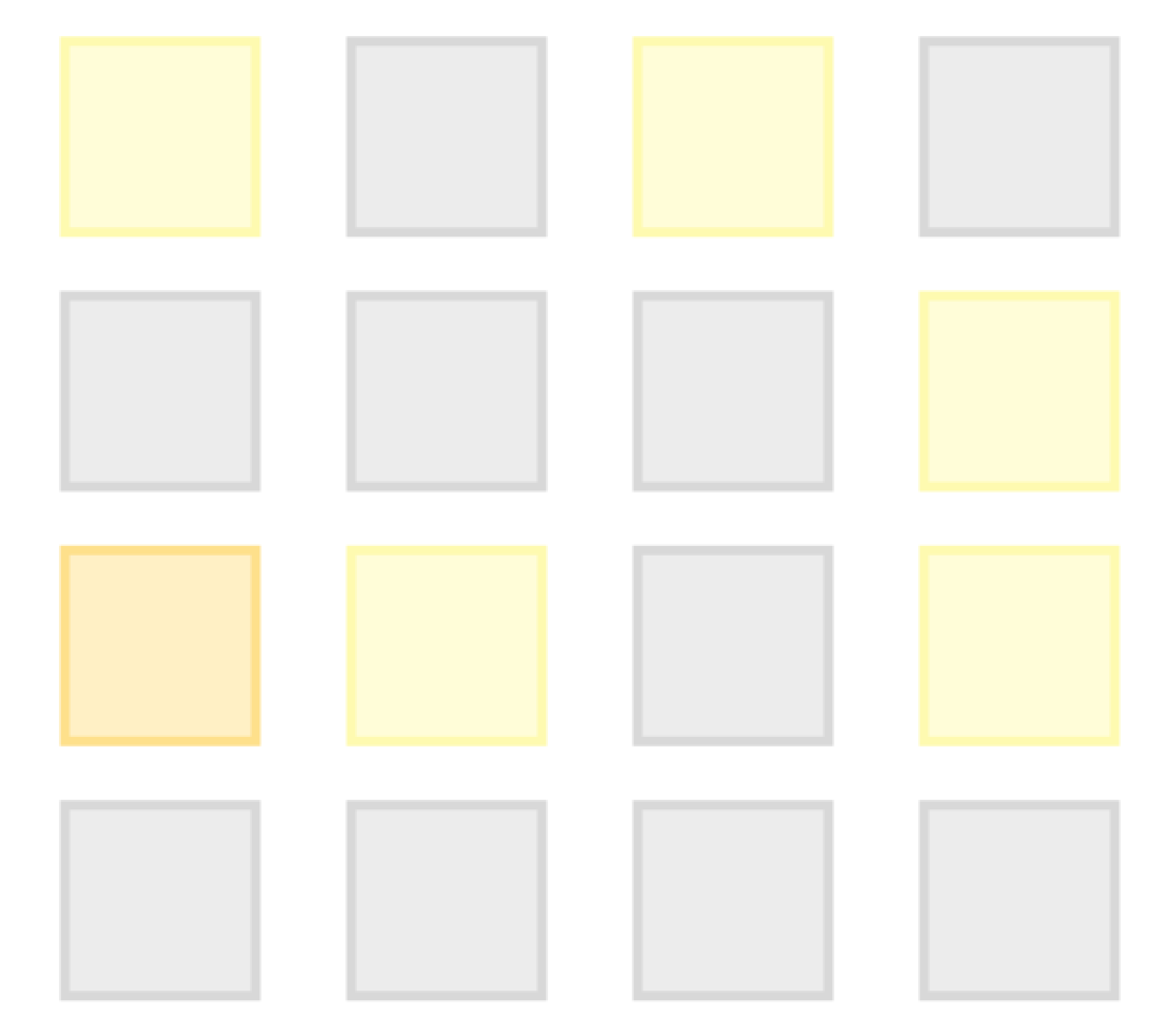}}
\caption{Topical behavior. (a) data points in high dimensional space; (b) cluster centers (topics) after dimension reduction; (c) topics after homogeneous mapping; (d) topical metrics for an entity}
\label{fig-TP}
\end{figure*}

\begin{figure*}[tb]
  \centering
  \includegraphics[scale=0.12]{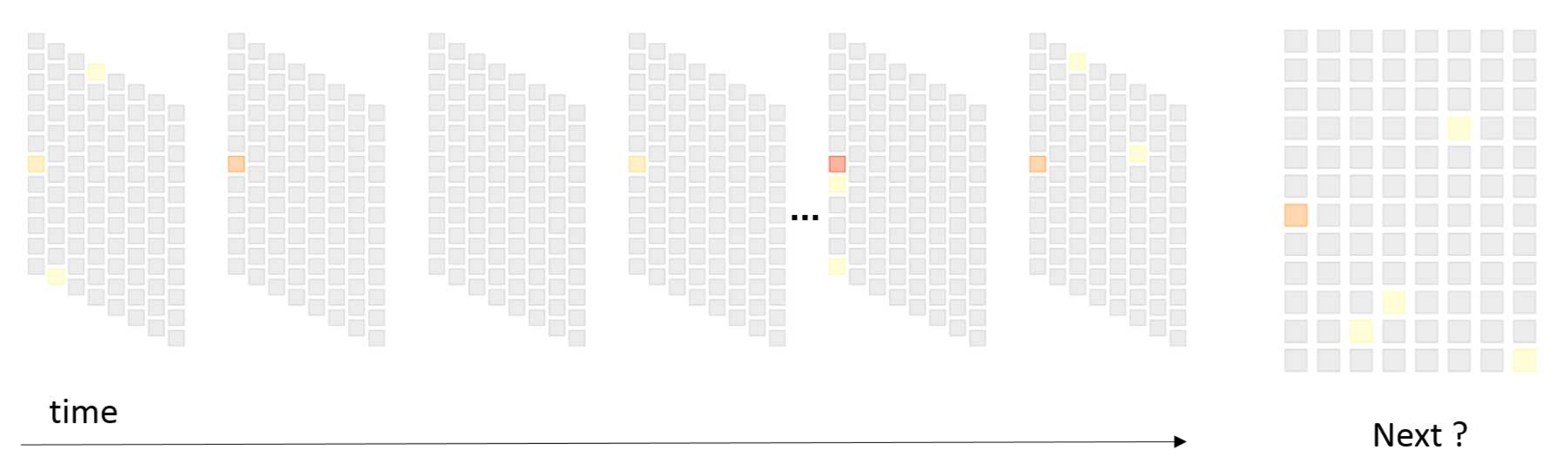}
\caption{Predicting the topical behavior of a single entity}
\label{fig-TR}
\end{figure*}

The main contribution is not combinining CNN and RNN. Instead, this work is the first attempt in applying the CNN and the RNN to predict topically summarized behavior. It is innovative in using both the spatial topic-to-topic relationship and the topic-over-time temporal trend for such task. The remaining of the paper is organized as follows. The proposed learning architectures to explore the temporal and spatial relationship on topics are presented in Section~\ref{sec-method}. Experiment setups and performance evaluation are covered in Section~\ref{sec-perform}. The result and comparison are discussed in Section~\ref{sec-discuss}. Finally, the conclusion and extension applications are provided in Section~\ref{sec-future}.

\section{Method} \label{sec-method}

\subsection{Topical Metrics}

To keep the behavior prediction within a trackable scope, we summarize the input behavioral data into topics. Starting from the activity log of all entities in the system, the descriptor vector of each activity entry is generated in a high dimension space. Clustering algorithm such as kmeans or latent latent Dirichlet allocation (LDA) \cite{Blei03} finds the topics (cluster centers) in this space. For each entity, the vectorized log entries are summarized on these topics to form quantitative metrics. For example, the topical volume over topic $t$ of entity $e$ can be measured as
\begin{equation} \label{eq-vol}
V^{(B_{e,T})}_t = \log(\sum_{a\in B_{e,T}}r_a+1)),
\end{equation}
where $r_a$ is the relevancy for activity $a$ to topic $t$, and $B$ is the set of activities defined by the unique content documents of all activities logged within the time period $T$. Similarly, to capture the trend in each topic, the topical drift between time periods $T_1$ and $T_2$ over topic $t$ for the same entity $e$ can be measured as
\begin{equation} \label{eq-risk}
\begin{aligned}
R^{(B_{e,T_1},B_{e,T_2})}_t ={} & \log(\sum_{a\in B_{e,T_2}}r_a+1) \\
& - \log(\sum_{a\in B_{e,T_1}}r_a+1).
\end{aligned}
\end{equation}

There are other types of topical quantifiers based on different business needs, such as topical risk, topical cost, or topical 'authority'-ness. Our goal in this paper is to track and predict the topical metrics over time, while leaving the topical metrics open to the downstream application. In the following sections, the topical volume similar to Eq~\ref{eq-vol} is applied without further mentioning.

\begin{figure*}[!h]
  \centering
  \subfloat[MLP]{\includegraphics[scale=0.082]{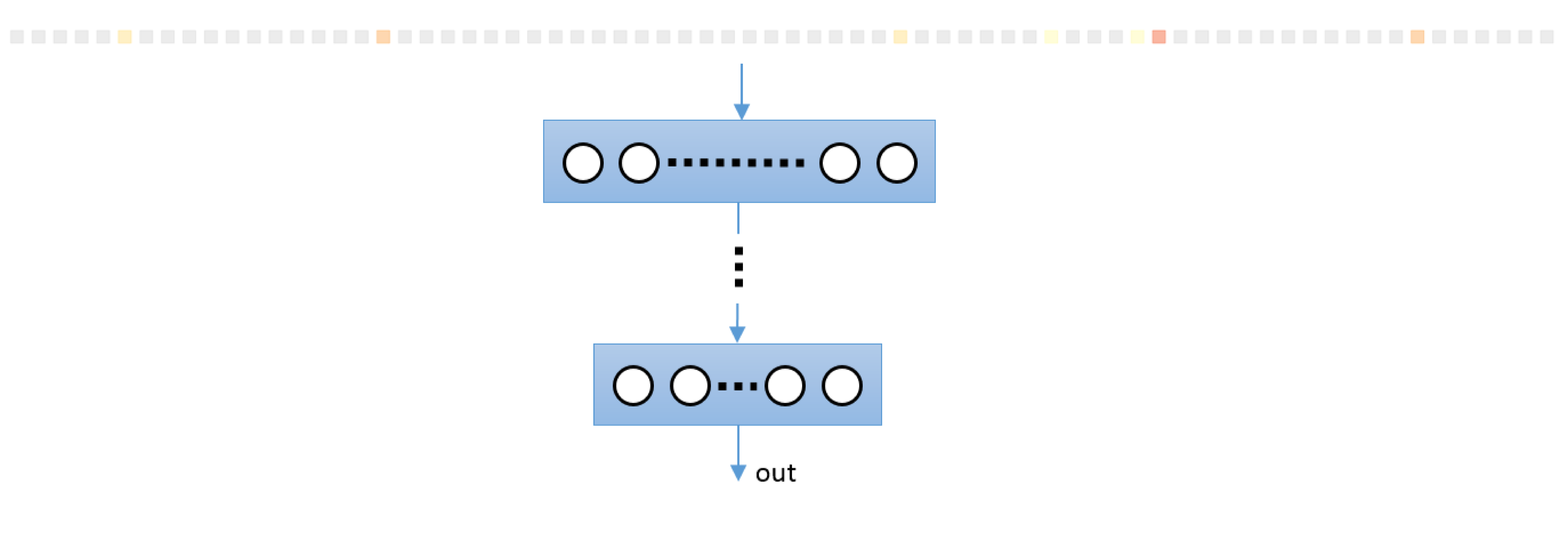}} \quad \vrule \quad
  \subfloat[TDRN]{\includegraphics[scale=0.082]{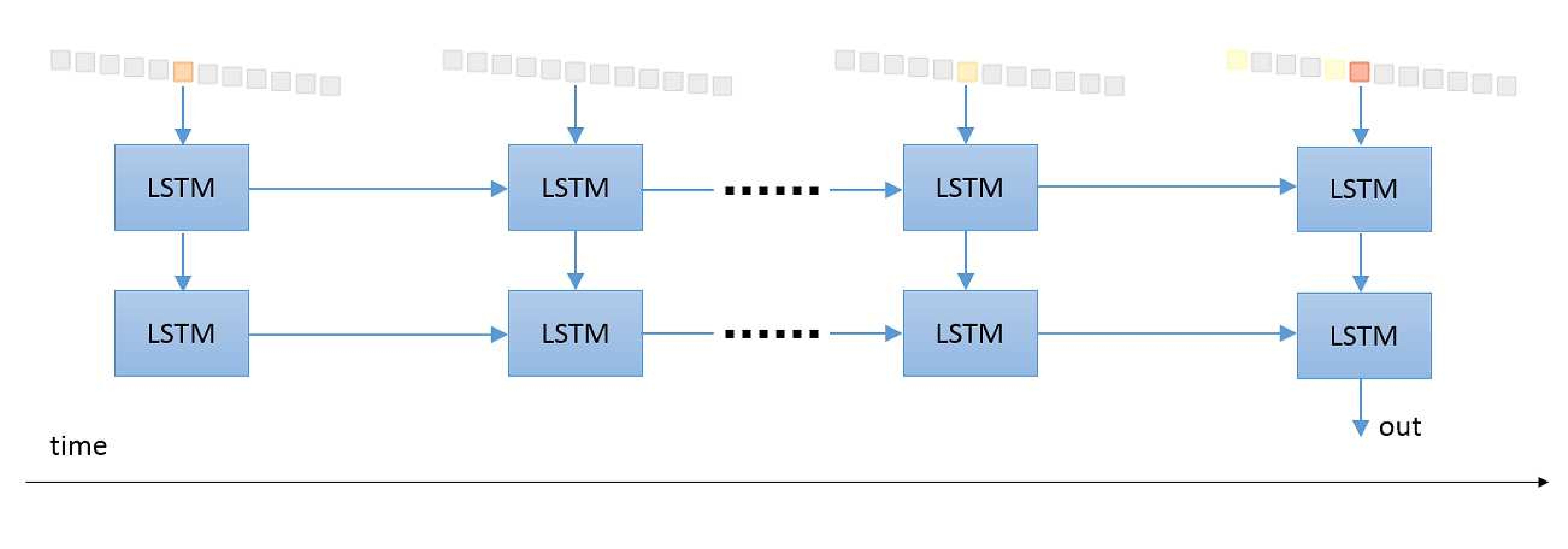}}
  \\
  \subfloat[LRCN]{\includegraphics[scale=0.082]{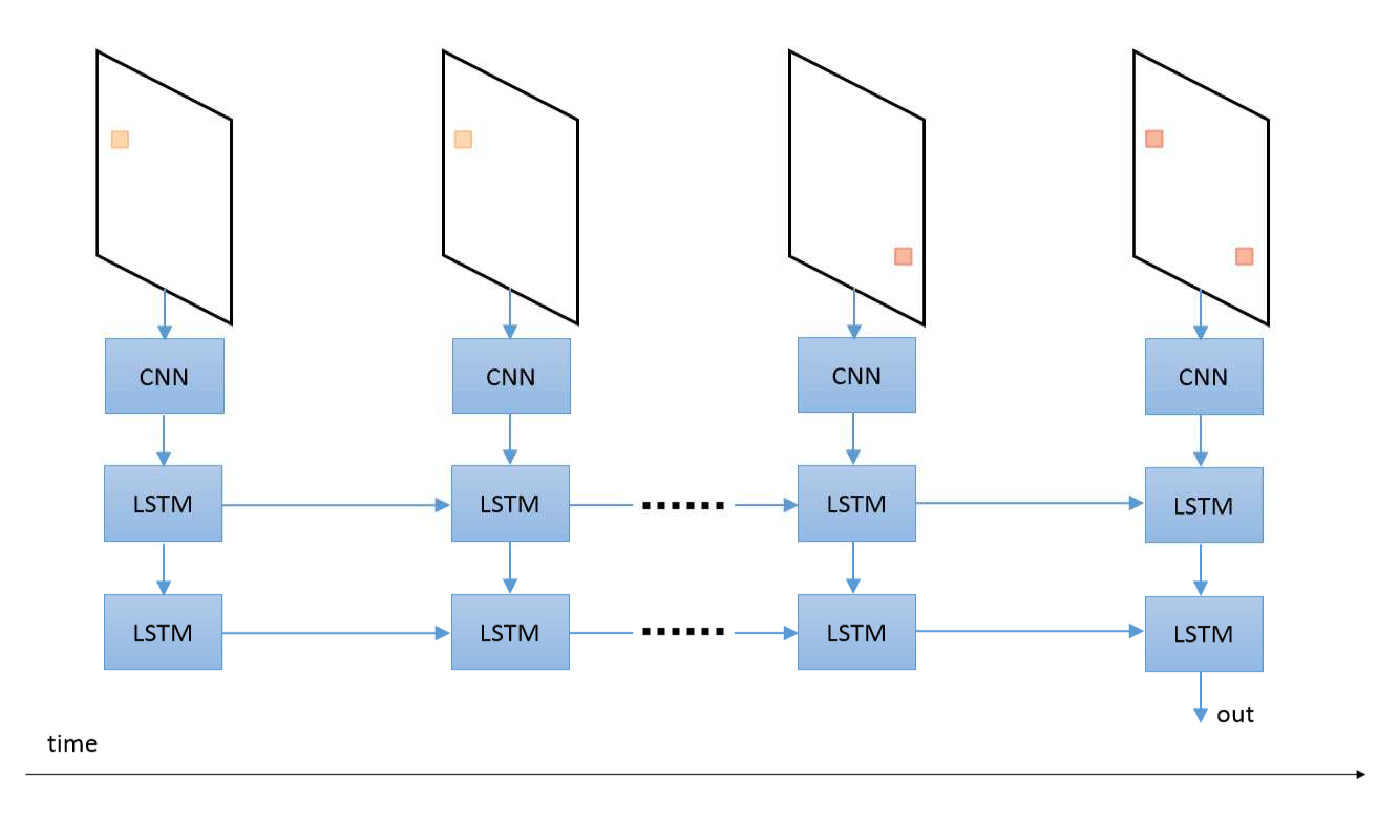}} \quad \vrule \quad
  \subfloat[SCCN]{\includegraphics[scale=0.082]{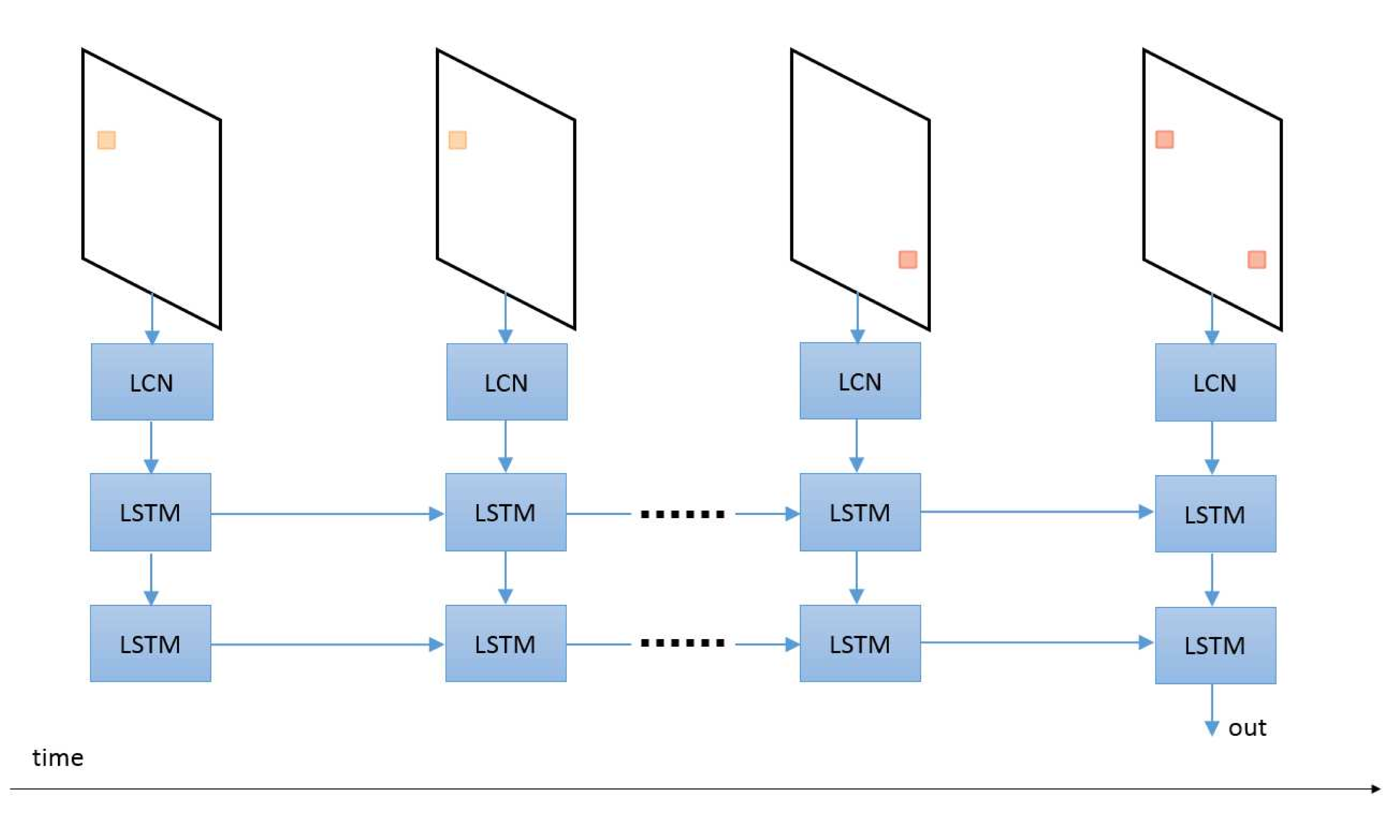}}
\caption{Different learning architecture for topical behavior prediction. (a) MLP tracks all topical behavior of any entity over all time periods as cascaded 1D vector; (b) TDRN uses the first layer of LSTMs to track the 1D topical behavior within each time period, then the second layer LSTM to track the states of the first layer LSTMs; (c) LRCNs are similar to TDRN while the behavioral relationship between topics is abstracted 2-dimensional and being learned by the CNN, before being tracked by the two LSTM layers; (d) SCCN replaces the CNNs in LRCN with the LCNs, whose vocabulary filters are not shared across different positions.}
\label{fig-deep}
\end{figure*}

\subsection{Spatial Reduction over Topics}\label{sec-srt}

To better explored the intra-topic relationship in the behavioral data, we want to capture the co-occurrence detail between any pair of topics. Furthermore, we want to learn the detail in the order of the distance between the topic pair - the co-occurrence means more when the two topics are closer to each other. 

CNN \cite{lawrence97} is a type of neural network that can learn the relationship at the finest pixel level, summarize (or pool) at the next level of granularity, then goes on to learn the element-wise relationship at that next level, until the top level. CNN has shown great success in classifying images \cite{krizhevsky12}, text \cite{kalchbrenner14}, videos \cite{karpathy14}, as well as other tasks in machine learning. It has the capability we want to learn the topical relationship in various detail at the same time. 

However, the CNN generally extracts pixel-wise relationship in the 2D or the 3D space, where the data points are distributed evenly. It cannot handle the topic-wise relationship directly in the word vector space. Therefore, the topical metrics need to be mapped into a space that the CNN can be applied. The same concept has been adopted for sentiment classification in \cite{yang16}, where the $n$-dimensional word embedding is applied to the $k$-word tweet and generates the $n$ by $k$ input array for the CNN.

In order to arrange the topical metrics in the format similar to the pixels in an image, the topical metrics need to go through the following two steps.
\begin{enumerate}
  \item Dimension reduction: this step maps the topical metrics into a 2-dimensional (2D) or 3-dimensional (3D) space, while maintaining the spatial relationship on topics according to different criteria of the dimension reduction algorithms. Some popular methods include principal component analysis (PCA) \cite{Pearson01}, multi dimensional scaling (MDS) \cite{Torgerson52}, and stochastic neighbor embedding (t-SNE) \cite{Van08}.
  \item Homogeneous mapping: on the visualization space that the CNN can digest, the topical metrics also need to be placed evenly. As in the dimension reduction step, the spatial relationship among the topics needs to be maintained after the mapping. One way to achieve this goal is the split-diffuse (SD) algorithm \cite{Su16}.
\end{enumerate}

The whole topical reduction process is illustrated in Figure~\ref{fig-TP}. In Figure~\ref{fig-TP}(a) shows all data points (vectorized log entries) in a high dimensional space, where color mean different topics. After the dimension reduction step, the cluster centers, representing the topics, are placed onto a visible space as in Figure~\ref{fig-TP}(b). Figure~\ref{fig-TP}(c) shows the homogeneously mapped topics. The topical metrics for an entity can be rendered like Figure~\ref{fig-TP}(d) to the CNN, or other learning architecture which exploits the spatial relationship.

When utilizing both the temporal and the spatial relationship among topics, the topical behavior prediction task for a single entity is illustrated in Figure~\ref{fig-TR}. The strategy on how to utilize the relationship and how to evaluate the performance is covered in the following.

\begin{table*}[h]
\caption{Performance of various learning architectures based on RLE. $\Delta$ means the gain versus MLP.}
\label{table-pe}
\vskip 0.15in
\begin{center}
\begin{sc}
\begin{tabular}{c|ccccc}
\hline
Architecture & $RLE$ & $\Delta RLE$ & $RLE_5$ & $\Delta RLE_5$& $RLE_1$ \\
\hline
MLP    & $0.1409$ & $-$ & $0.1364$ & $-$ & $0.1363$ \\
TDRN   & $0.1192$ & $15.44\%$ & $0.1162$ & $15.62\%$ & $0.1182$ \\
LRCN   & $0.1128$ & $19.92\%$ & $0.1083$ & $20.53\%$ & $0.1079$ \\
SCCN   & $0.1129$ & $19.85\%$ & $0.1096$ & $19.63\%$ & $0.1095$ \\
LRCNM  & $0.1125$ & $20.13\%$ & $0.1073$ & $21.28\%$ & $0.1076$ \\
SCCNM  & $0.1116$ & $20.79\%$ & $0.1083$ & $20.59\%$ & $0.1080$ \\
\hline
\end{tabular}
\end{sc}
\end{center}
\end{table*}

\begin{table*}[h]
\caption{Performance of various learning architectures based on R2LE. $\Delta$ means the gain versus MLP.}
\label{table-pe2}
\vskip 0.15in
\begin{center}
\begin{sc}
\begin{tabular}{c|ccccc}
\hline
Architecture & $R2LE$ & $\Delta R2LE$ & $R2LE_5$ & $\Delta R2LE_5$ & $R2LE_1$ \\
\hline
MLP    & $0.3225$ & $-$ & $0.3105$ & $-$ & $0.3100$ \\
TDRN   & $0.2858$ & $11.39\%$ & $0.2836$ & $10.03\%$ & $0.2809$ \\
LRCN   & $0.2782$ & $13.73\%$ & $0.2649$ & $14.70\%$ & $0.2617$ \\
SCCN   & $0.2746$ & $14.84\%$ & $0.2664$ & $14.20\%$ & $0.2667$ \\
LRCNM  & $0.2813$ & $12.77\%$ & $0.2586$ & $16.72\%$ & $0.2559$ \\
SCCNM  & $0.2737$ & $15.12\%$ & $0.2624$ & $15.50\%$ & $0.2623$ \\
\hline
\end{tabular}
\end{sc}
\end{center}
\end{table*}

\subsection{Temporal and Spatial Learning}

We adopt various DNN architectures to study how the temporal and the spatial information can help learning and predicting the topical metrics. 

MLP with layers of neurons in Figure~\ref{fig-deep}(a) serves as the benchmark. In the MLP case, the topical metrics over different time periods are cascaded into one single 1-dimensional (1D) vector for each sample. The number of neurons gradually reduces over layers, with the output layer matching the size of the response space which is the number of topics.

In utilizing the temporal relationship among topics, we use one layer of LSTMs to track the topical metrics as a sequence for each time period, and then another layer of LSTM to track the output states of the LSTMs from the previous layer. The whole framework is called the time distributed recurrent network (TDRN), as shown in Figure~\ref{fig-deep}(b).

The long-term recurrent convolutional networks (LRCN) is a family of architectures introduced to generate captions for videos \cite{donahue15}. It combines the convolutional layers with the long-range temporal recursion. We employ the LRCN to exploit both the temporal and the spatial relationship among topics. In our implementation, the uniformly distributed topical metrics from Subsection~\ref{sec-srt} are spatially learned by CNN, then the LSTM layers track the variation among the spatial patterns over time, as in Figure~\ref{fig-deep}(c).

However, the CNNs are computationally expensive. LCN is another option to explore spatially while being time efficient. In the context of speaker recognition, LCN performs comparably well versus CNN \cite{chen12}. In the framework called spatially connected convolutional networks (SCCN) as in Figure~\ref{fig-deep}(d), the convolutional units in LRCN are replaced by the locally connected units. The locally connected units do not share the trained weights between different position. Instead, the same set of weights is applied to the same position of different samples. The regulation is more effective on the locally customized  patch dictionaries in the LCN, compared to that on a global dictionary in the CNN.

In our experiments, we also consider the multi-resolution approach to learn the spatial and temporal relationship. For the LRCN case, four subsystems as Figure~\ref{fig-deep}(c) with different vocabulary filter sizes are fed with the same tensor input. In the final stage, the outputs from all four subsystems are collected to generate the response metrics. The same multi-resolution approach is carried over to the SCCN as well.

\section{Performance}  \label{sec-perform}

\subsection{Measuring Losses}

The common way to measure the prediction result is the mean squared error (MSE) between the predicted values and true values, over all topical metrics of all samples. However, the cost of the decision based on the behavioral prediction result usually is not equally weighted across each predicted value. For example, in predicting the trending or risky topic, the cost of missed future trend is higher than the cost of false positives. Therefore, we use the following weighted losses to evaluate the behavioral predictions.

The risk loss error (RLE) weighs the squared error linearly according to the true target value, such as the $V^{(B_{e,T})}_t$ in Eq~\ref{eq-vol}.
\begin{equation} \label{eq-rle}
 RLE = \frac{1}{|\mathcal{V}|} \sum_{\forall v \in \mathcal{V}} {v(\hat{v}-v)^2},
\end{equation}
where $\mathcal{V}$ is the set of all target values, which are the non-negative and scaled topical metrics. We further emphasize the missed prediction error in another loss called risk square loss error (R2LE) defined as
\begin{equation} \label{eq-r2le}
 R2LE = \frac{1}{|\mathcal{V}|} \sum_{\forall v \in \mathcal{V}} {v^2(\hat{v}-v)^2}.
\end{equation}
The entity usually are active on few topics out of many. To better regulate the learning, 
both RLE and R2LE encourage the learning architecture to predict boldly on high potential topics, instead of passively assigning zeros to optimize MSE.

We want to leave the metric open according to the downstream applications. In our example of topical risk, RLE and R2LE are designed to penalize the prediction error on highly active topics, such as the orange grids in Figure~\ref{fig-TR}. Meanwhile, the error with low-activity topics is tolerated. The same metric also works for predicting the topical sales volume on a ecommerce site. However, to predict the shopper's change in flavor, the error can be measured on the binary difference of each shopping topic.

\begin{figure*}[tb]
  \centering
  \subfloat[$RLE$ on training]{\includegraphics[scale=0.15]{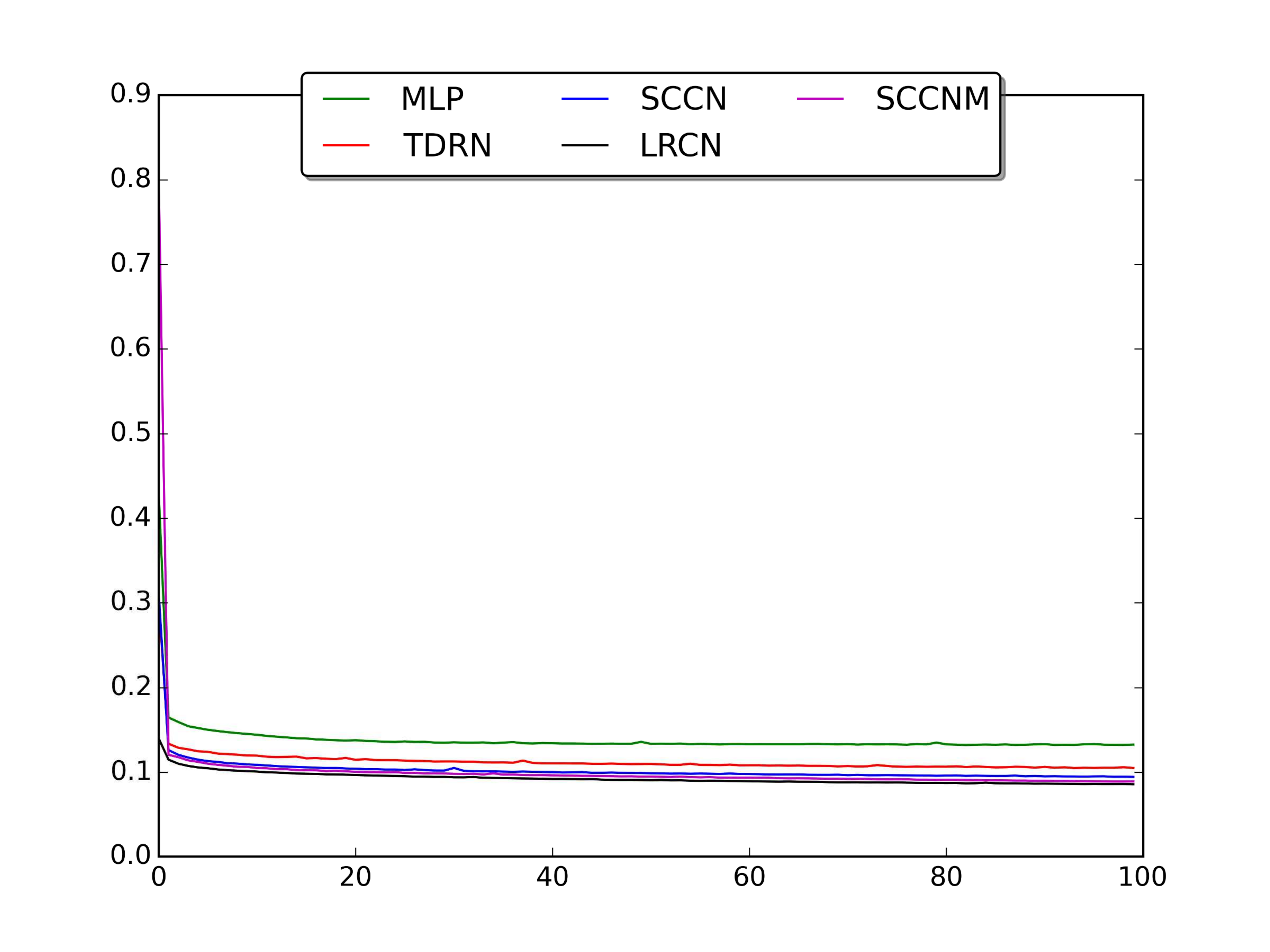}} 
  \subfloat[$RLE$ on validation]{\includegraphics[scale=0.15]{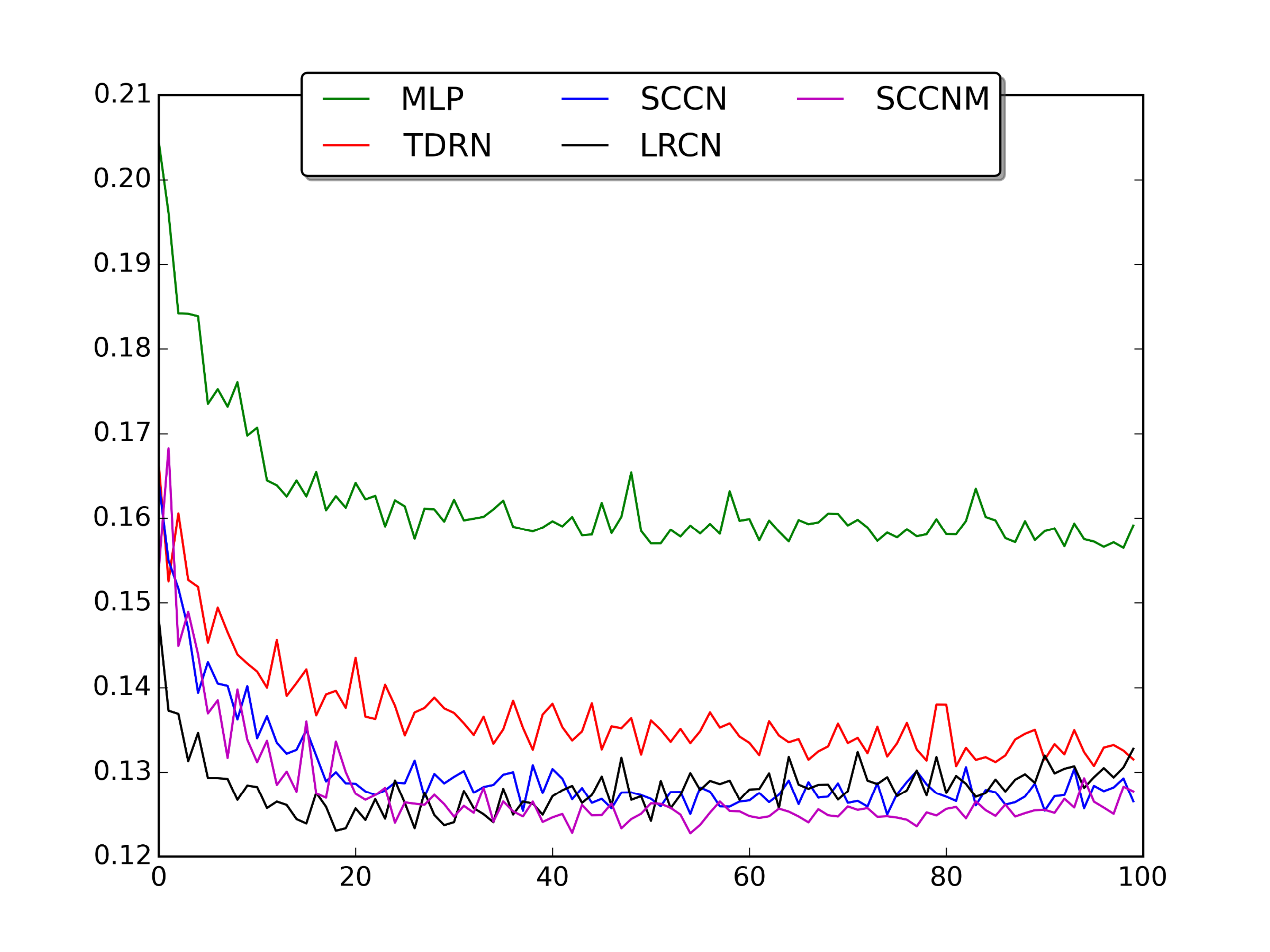}} 
  \subfloat[$RLE$ on testing]{\includegraphics[scale=0.15]{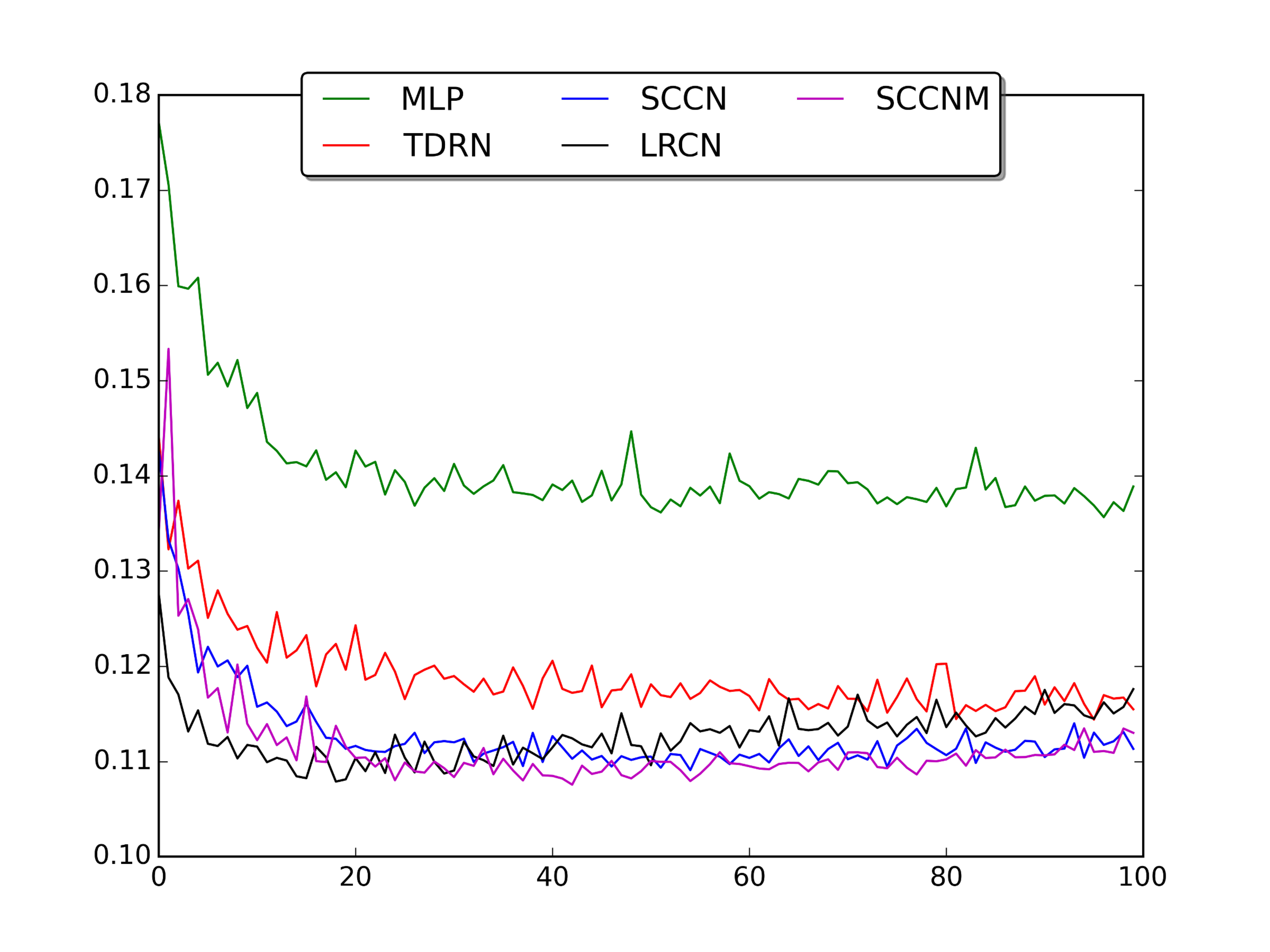}}
  \\
  \subfloat[$R2LE$ on training]{\includegraphics[scale=0.15]{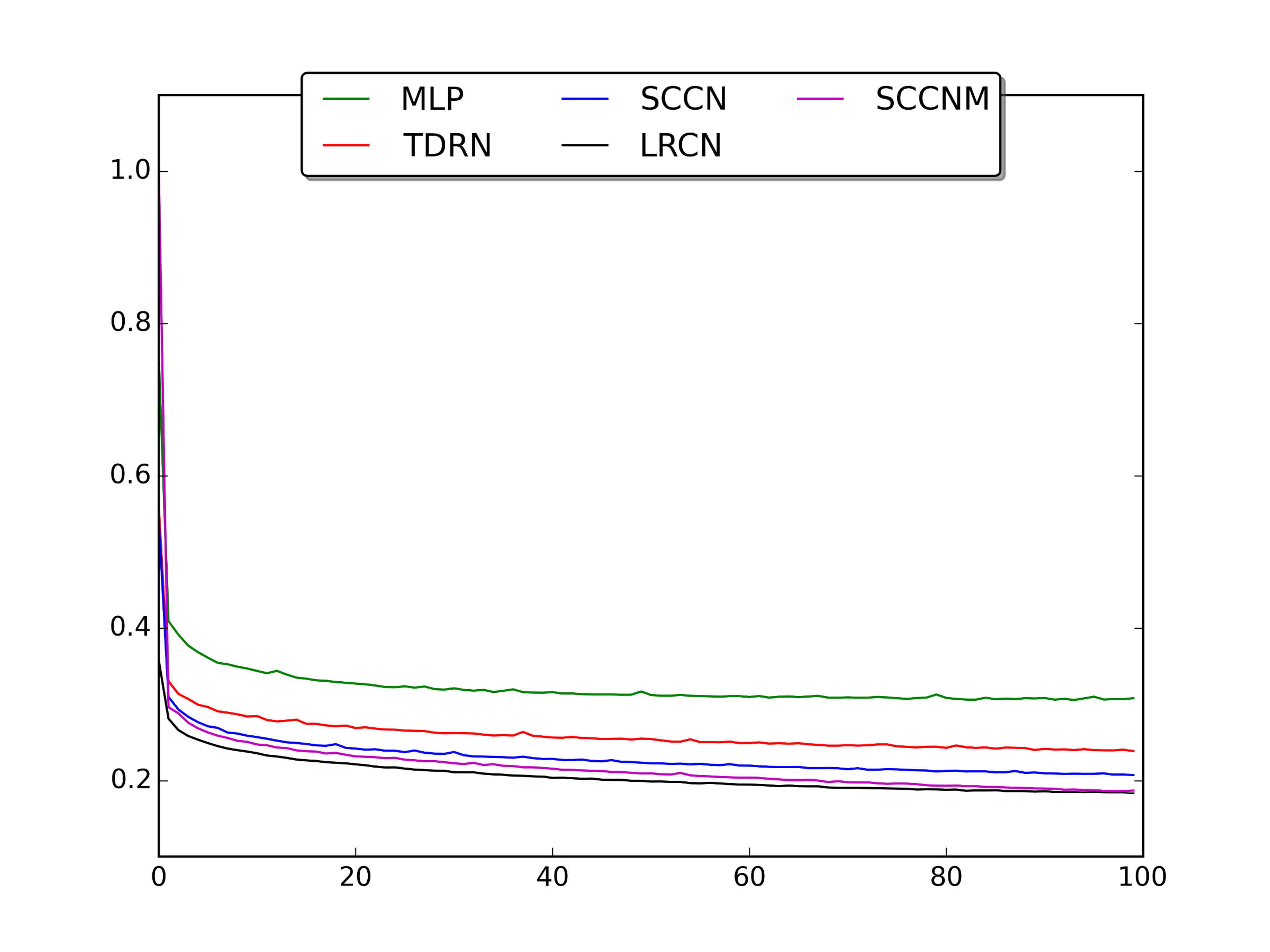}}
  \subfloat[$R2LE$ on validation]{\includegraphics[scale=0.15]{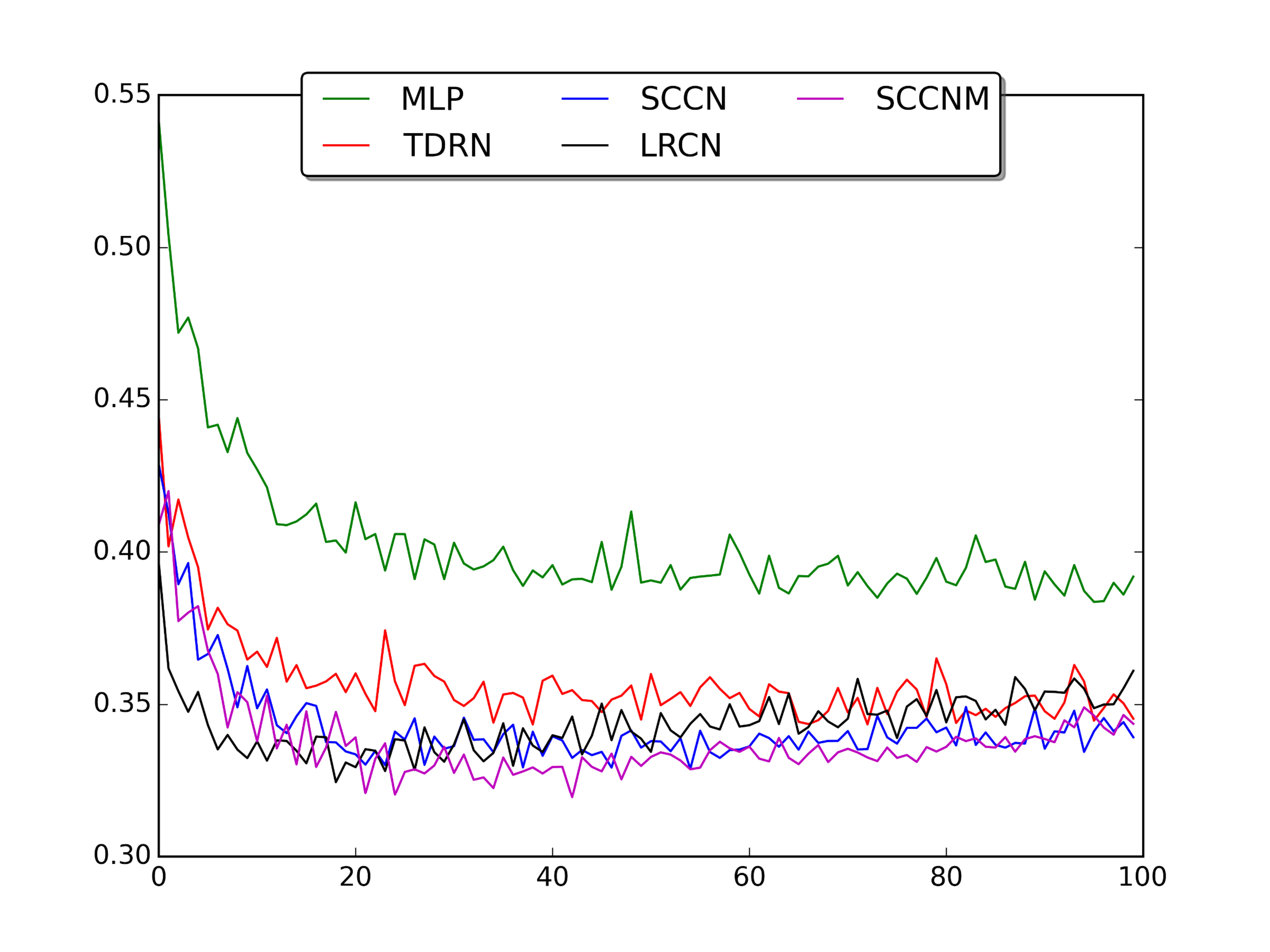}}
  \subfloat[$R2LE$ on testing]{\includegraphics[scale=0.15]{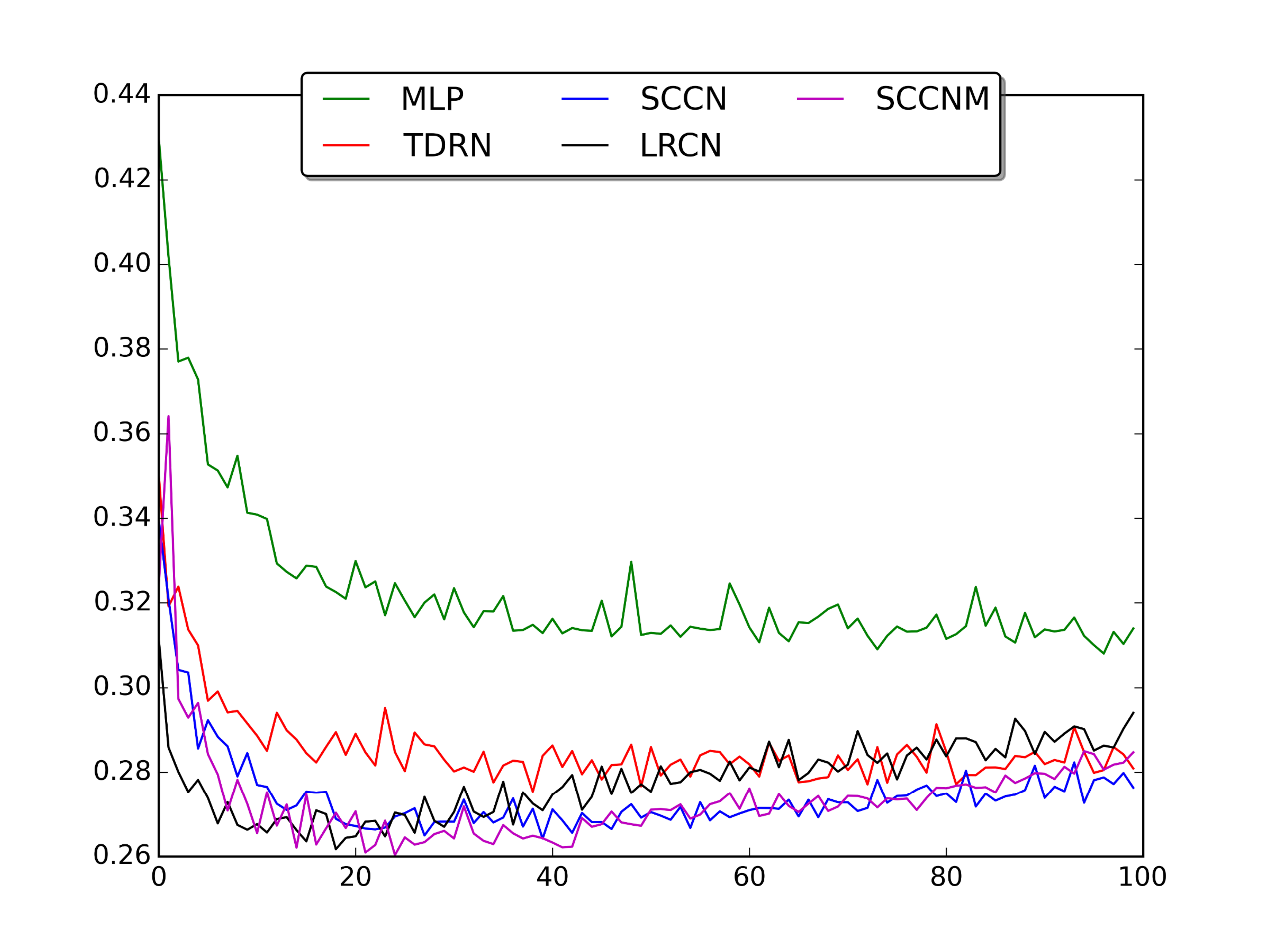}}
\caption{Performance on various learning architectures; x-axis indicates the number of epochs.}
\label{fig-R}
\end{figure*}

\subsection{Experiment Setup}

The data set comes from $151,783,806$ entries in the activity logs of $98,881$ samples. Each entry contains the identifier of the acting entity, the access timestamp, and the meta information about the accessed resource. The meta information of each entry is punctuated into words and vectorized, called the content document. The the content documents of the first six months are used to train the LDA and extract $96$ topics. The trained LDA is then fixed and provides the relevancy for each log entry on each topic.

A sample comes from a single entity. Over each time period, the topical relevancy of each sample is quantified by Equation (\ref{eq-vol}) for every topic. Figure~\ref{fig-TR} illustrated what a sample looks like with the help from the Topic Grids \cite{Su16}. Each sample contains the topical relevancy on each of all $96$ topics, from all historical periods up to the current period. Each sample also contains the prediction target which is the topical relevancy of these $96$ topics in the next (future) period. 

The samples are split into $69,407$ training samples, $7,712$ validation samples, and $21,762$ testing samples. The prediction target values in the testing samples are from the time periods that are later then all the training time periods. That means, with May being the last time period to train the architecture, the earliest possible test set prediction happens in June of the same year. However, overlaps in the time periods where the historical topical metrics were generated are allowed between training data and testing data. This setup simulates the behavioral prediction for the future.

The learning architectures in our implementation are built with Keras \cite{chollet15keras} with Theano \cite{Theano} backend. Both dropout \cite{hinton12} and $L_2$ regulation \cite{ng04} are applied to all the architectures to keep the learned models generalized enough.

\subsection{Result}

Tables~\ref{table-pe} and~\ref{table-pe2} show the experimental result on loss metrics $RLE$ and $R2LE$ over  all the learning architectures discussed above. The epochs when the validation data has the best five $RLE$s are chosen to form the $RLE_5$ by averaging the corresponding $RLE$s from the testing data. $R2LE_5$ is derived in the same fashion. The prediction gain comparing against the benchmark MLP result is denoted with the $\Delta$ prefix. LRCNM and SCCNM are the multi-resolution setups for LRCN and SCCN, respectively.

The TDRN explores only the temporal relationship with LSTM over topics. It provides prediction gain ranging from $11.32\%$ to $16.73\%$ depending on the loss metric and the evaluation scenario. The LRCN explores both temporal relationship and spatial relationship over topics. It tracks the temporal relationship with layers of LSTMs, exactly the same approach used in TDRN. The additional spatial information among topics tracked by CNN further improve the prediction gain from $13.73\%$ to $19.92\%$. Replacing the CNN spatial tracking with the LCNs, the SCCN provides a comparable $14.20\%$ to $19.85\%$ prediction gain to the LRCN.

In the multi-resolution setups, four subsystems with vocabulary filters sizing $4$-, $9$-, $16$- and $25$-topics are combined to learn the behavioral metrics. The LRCNM yields a wider range of gain from $12.77\%$ to $21.28\%$, which can be substantial when guided properly by the validation data. The SCCNM, on the other hand, provides a more stable gain ranging from $15.12\%$ to $20.79\%$.

In our experiments, both RLE and R2LE are measured at the level of the predicted values, not at the sample level. The benefit on this approach is that the loss is accurately measured at the level where the prediction is made. The detail learning curves for all architectures are shown in Figure~\ref{fig-R}. To match the other learning architectures in terms of learning capability, the benchmark MLP is implemented with five layers of neurons, including input and output layers, and three dense neuron layers in absence of the LSTM and spatial learning layers in other architectures in Figure~\ref{fig-deep}.

The $L_2$ regulation is applied to all learning architectures shortly after the input. For the MLP, it is applied to the first dense neuron layer after the input neurons. For the TDRN, the $L_2$ regulation works on the time distributed dense layer after the input at each time period. In a similar fashion, $L_2$ is applied to the LRCN and the SCCN at the CNN and the LCN layer of each time period, respectively. Each subsystem in the LRCNM and the SCCNM architectures also employ the same $L_2$ as the corresponding single resolution architecture. The dropouts are used in the same layers in each architecture.

\begin{table*}[t]
\caption{Increase in architecture size on multi-resolution setups. Number of filters is regarding to the filters used in each single spatially-convoluted network (CNN or LCN).}
\label{table-mr}
\vskip 0.15in
\begin{center}
\begin{sc}
\begin{tabular}{ccr|ccr}
\hline
Architecture & $\#$ Filters & $\#$ Neurons & Architecture & $\#$ Filters & $\#$ Neurons \\
\hline
LRCN & $16$ & $480,160$ & LRCNM & $16$ & $1,798,144$ \\
LRCN & $32$ & $848,960$ & LRCNM & $32$ & $3,113,888$ \\
LRCN & $64$ & $1,586,560$ & LRCNM & $64$ & $5,745,376$ \\
SCCN & $16$ & $489,600$ & SCCNM & $16$ & $1,838,528$ \\
SCCN & $32$ & $867,840$ & SCCNM & $32$ & $3,194,656$ \\
SCCN & $64$ & $1,624,320$ & SCCNM & $64$ & $5,906,912$ \\
\hline
\end{tabular}
\end{sc}
\end{center}
\end{table*}

\section{Discussion}  \label{sec-discuss}

\subsection{Spatial Gain}

The spatial gain comes from asking the CNN neurons or the LCN neurons to learn the pixel-wise relationship locally, and summarize the learned knowledge across layers, whose definition on 'localness' becomes larger and larger, up to the whole image. However, the topical metrics fed into the convolutional neurons are not truly pixels of a natural image. Instead, the relationship between topics are reduced significantly to map the topical metrics uniformly like the pixels.

When being generated in a high dimensional space from the sampled data points by the generative models \cite{bishop07}, the topics close to each other indicate that their context may have overlap. It also means that a data point is more likely to belong to the topics which are nearby spatially than the topics that are not. If the topically related metrics are still close by after the heavy reduction into pixels, the convolutional neurons can systemically learn the pixel-wise relationship into meaningful vocabulary filters.

Even after the series of reductions, the result from both the LCRN and the SCCN confirms the spatial gain when compared against the TDRN. The redefined neighborhood at the pixel level provides an overall $1.21\%$ to $3.41\%$ on top of the temporal gain. 

\subsection{Temporal Gain}

The temporal gain is observed by inspecting the prediction difference between the MLP and the TDRN. Depending on the loss metrics being evaluated, the temporal gain ranges from $12.52\%$ to $16.51\%$ overall. In our experiments, both the temporal and the spatial gains show up on $RLE$ and $R2LE$ across the topical metric on the quantity defined in Eq~\ref{eq-vol}. Different loss metrics on different topical metrics may vary the gain range. The learning architectures can be another factor, while efforts are made to keep the architectures comparable to each other.

\subsection{CNN versus LCN}

As the convolutional neurons try to pick up the recurrent founding components spatially, the difference between the CNN and the LCN lies in whether the learned vocabulary filters are shared across different positions when filtering the pixels. The result in Tables~\ref{table-pe} and~\ref{table-pe2} show similar performance between the CNN and the LCN. 

Theoretically, the patterns on topical metrics carry the information about how the specific topics within the filter are related to each other. If one considers that each topic is unique in its own contextual meaning, then each relationship is also uniquely defined, for example, the relationship between the topic 'baseball' and the topic 'reggae music'. The locally connected convolutional neurons make more sense when each topic is uniquely presented in a single position, which is the case of ours. On the other hand, the neurons in the CNN are designed to keep track of the spatially recurrent patterns, for example, a wheel at different positions of the images.

As shown in Figure~\ref{fig-R}, the CNN quickly learns the topical relationship with the spatially shared neurons. However, it begins to overfit even with the exact same regulations used in the LCN. At this point, the spatial constraints enforced in the LCN can be seen as a form of spatial regulation for the convolutional neurons. With the better regulated learning architecture, we do not need to reserve a portion of the training data to validate and probe the best trained states. Thus LCN can better utilize the data in our experiments. Nevertheless, the run time for the LCRN is between $1.5$ to $3$ times that for the SCCN. This is similar to the observation in \cite{chen12} where the CNN and the LCN are trained to recognize the speaker.

\subsection{Multi-resolution}

In a multi-resolution learning architecture, each subsystem uses a fixed size of filter bank for its spatial convolutional neurons. For SCCNM, the multi-resolution prediction gain on top of the single-resolution SCCN ranges from $0.28\%$ to $1.30\%$. The multi-resolution setup worsens the overfitting for the LRCN. While the result guided by the validation data gains from $0.52\%$ to $1.65\%$, the overall performance is not improved.

The combined subsystems in the multi-resolution setup also significantly increase the model size (Table~\ref{table-mr}) and the computation time. In our experiments, the better regulated SCCN is more appropriate for the multi-resolution type of learning.

\section{Conclusion and Future Work}  \label{sec-future}

\subsection{Conclusion}

In this paper, we formalize the behavior prediction over topics, utilizing both the topic-wise relationship and the topical variation in time. Several state-of-the-art learning architectures are evaluated under the proposed behavior prediction framework. The experiments are designed to incrementally add the temporal and the spatial information into the learning. Our result shows consistent gains on learning the temporal detail and the spatial detail.

We present SCCN, a type of learning architecture to explore deeply in both the spatial and the temporal domains. The SCCN provides comparable result to another type of architecture, LRCN, in the loss metrics tested. It is faster to train and to make prediction. Meanwhile, it is better regulated, making it more suitable for larger scale behavioral learning. 

\subsection{Extension}

There are several aspects that may be further investigated under the proposed framework. 

Toward the homogeneous placement of the topics, it is possible that the two dimension reduction steps in Section~\ref{sec-srt} can be combined into one in a unified approach, by adding the condition of homogeneous placement in the destination space as the criteria during dimension reduction. If exists, the unified approach may further improve the prediction performance. However, since different dimension reduction algorithms have different reduction objectives, the improvement on such unified approach may also depend on its use cases.

Each of the components in the learning architectures can be further replaced by its alternatives or variants. For example, it is also possible to use other types of recurrent units such as the GRU \cite{chung14} to track time, instead of the LSTM in this paper.

On the application side, the proposed framework can predict topical behavior for any logs with large amount of activities, from which the topics can be extracted. The log can be on network activities, financial transactions, social media posts, visited places on car rides, among many others. It is not limited to the unstructured data input. In fact, it is also applicable to the structured data where topics or cluster centers can be formed.

\bibliographystyle{IEEEtran}
\bibliography{IEEEabrv,jsu2016}

\end{document}